\documentclass[10pt]{article}
\usepackage{arxiv}
\usepackage{cite}
\usepackage[utf8]{inputenc}
\usepackage[T1]{fontenc}
\usepackage[protrusion=true,expansion=false]{microtype}
\usepackage{amsmath,amssymb,amsfonts}
\usepackage{graphicx}
\usepackage{booktabs}
\usepackage{multirow}
\usepackage{xcolor}
\usepackage{subcaption}
\usepackage{url}
\usepackage[colorlinks=true,linkcolor=blue,citecolor=blue,urlcolor=blue]{hyperref}

\newcommand{\orcidicon}[1]{\href{https://orcid.org/#1}{\includegraphics[width=0.32cm]{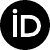}}}

\newcommand{\stmoe}{\textsc{ST-MoE}}
\newcommand{\dmodel}{d_\text{model}}
\newcommand{\dspace}{d_\text{space}}
\newcommand{\topk}{\text{top-}k}

\newcommand{\eg}{\textit{e.g.}}

\newcommand{\Paragraph}[1]{\vspace{0.5em}\noindent\textbf{#1}}

\title{Geometric Routing Enables Causal Expert Control \\
in Mixture of Experts}

\author{
  Ivan Ternovtsii \orcidicon{0009-0009-9267-8516}\thanks{This research was conducted as part of PhD studies at the Department of Software Systems, Faculty of Information Technologies, Uzhhorod National University. HengeBytes generously provided computational resources. Corresponding author: Ivan Ternovtsii (e-mail: ivan.ternovtsii@uzhnu.edu.ua).} \\
  Department of Software Systems, Uzhhorod National University\\
  Narodna sq. 3, Uzhhorod, Ukraine, 88000\\
  HengeBytes\\
  \texttt{ivan.ternovtsii@uzhnu.edu.ua} \\
  \And
  Yurii Bilak \orcidicon{0000-0001-5989-1643} \\
  Department of Software Systems, Uzhhorod National University\\
  Narodna sq. 3, Uzhhorod, Ukraine, 88000\\
}

\date{April 2026}

\begin{document}
\maketitle

% ============================================================
% ABSTRACT
% ============================================================
\begin{abstract}
Sparse Mixture-of-Experts (MoE) models scale parameters while fixing active computation per token, but the specialization of individual experts remains opaque.
In a companion paper \cite{equifinality2026} we showed that routing \emph{topology} is quality-neutral: five structurally different configurations converge to statistically equivalent language modeling quality.
Here we show that expert \emph{identity} is nonetheless causally meaningful: individual rank-1 experts are monosemantic by construction, and cosine-similarity routing in a low-dimensional metric space makes their specialization directly inspectable.

We present four lines of evidence.
First, projecting expert output vectors through the unembedding matrix yields a \textbf{Semantic Dictionary}: 15\% of experts are monosemantic specialists spanning 10 categories (temporal, geographic, cardinal, discourse, emotional, financial, military, scientific).
Second, routing exhibits a \textbf{frequency-to-syntax gradient}: early layers separate tokens by word frequency, deeper layers by syntactic class (Zipf-confound controls, all $p < 0.001$).
Third, \textbf{causal interventions} confirm these labels: steering toward a temporal expert's centroid increases P(temporal) by $+321\%$ (median across 44 prompts); suppressing a geographic expert drops P(geographic) by $-23\%$; rewriting an expert's output vector halves target-category probability, and effects compose additively across layers.
Fourth, the interventions are not unique to cosine routing: linear routers support comparable steering, but only cosine routing provides \textbf{geometric transparency}---expert specialization is readable directly from the centroid matrix.

MoE expert-level specialization is a first-class interpretability primitive: architecturally monosemantic, causally validated, and controllable at inference with zero overhead.
\end{abstract}


% ============================================================
% 1. INTRODUCTION
% ============================================================
\section{Introduction}
\label{sec:introduction}

Sparse Mixture-of-Experts (MoE) architectures scale model parameters while keeping per-token computation fixed, enabling models like Mixtral \cite{jiang2024mixtral} and DeepSeek-MoE \cite{dai2024deepseekmoe} to achieve competitive quality at reduced inference cost.
Yet the internal specialization of individual experts remains poorly understood: do experts develop meaningful, causal roles, or are they interchangeable components?

In a companion paper \cite{equifinality2026}, we showed that routing \emph{topology}---single-hop vs.\ multi-hop, shared vs.\ decoupled projections, cosine vs.\ linear routing---does not determine asymptotic language modeling quality.
Five structurally different routing configurations are statistically equivalent within a 1-PPL band (Two One-Sided Tests equivalence, $p < 0.05$, 15 runs across 3 seeds).
This equifinality result raises a natural question: if the \emph{path} through the expert pool doesn't matter, does the identity of individual experts matter?

This paper answers in the affirmative.
We show that individual rank-1 experts develop clear semantic specializations that are:
\begin{enumerate}
    \item \textbf{Discoverable}: Expert output vectors, projected through the unembedding matrix, reveal interpretable semantic functions---a technique we call the Semantic Dictionary (Section~\ref{sec:dictionary}).
    \item \textbf{Syntactically organized}: Routing separates tokens by syntactic class (not mere word frequency), with a depth gradient from coarse frequency sensitivity to fine-grained syntax (Section~\ref{sec:frequency}).
    \item \textbf{Context-sensitive}: Identical tokens route through 86\% different experts depending on semantic context, showing routing is genuinely semantic (Section~\ref{sec:polysemy}).
    \item \textbf{Causally validated}: Steering, suppression, and surgery interventions confirm that Semantic Dictionary labels are not merely descriptive but causally responsible for specific output distributions (Section~\ref{sec:interventions}).
    \item \textbf{Composable across layers}: Cross-layer expert pairs compose nearly additively, while same-layer pairs interfere---revealing a structural constraint on compositional control (Section~\ref{sec:composition}).
    \item \textbf{Routing-mechanism-general}: Linear routers support comparable steering effects, but cosine routing provides geometric transparency---expert functions are readable from centroids without running activation probes (Section~\ref{sec:cross_routing}).
\end{enumerate}

Together, these findings show that MoE experts are architecturally monosemantic, causally validated, and controllable at inference time.
Equifinality holds at the topology level; specialization holds at the expert level.

% ============================================================
% 2. BACKGROUND
% ============================================================
\section{Background}
\label{sec:background}

We summarize the architecture and training setup; full details are in \cite{equifinality2026}.

\Paragraph{Semantic Trajectory MoE (\stmoe{}).}
\stmoe{} is a pre-LayerNorm Transformer where each block replaces the dense feed-forward network (FFN) with a sparse multi-hop routing layer.
Given a token representation $h \in \mathbb{R}^{\dmodel}$, routing projects it into a low-dimensional coordinate space $\text{pos} = \text{normalize}(\text{proj}_\text{in}(h)) \in \mathbb{R}^{\dspace}$ ($\dspace = 64$), computes cosine similarity with $M = 1024$ learned expert centroids, and selects top-$K = 4$ experts via softmax with temperature $\tau = 30$.
Selected experts apply rank-1 MLP updates: $\Delta h = \sum_{i \in \topk} w_i \cdot W_{\text{up},i}\, \text{SiLU}(W_{\text{down},i}\, h)$.
After each of $H = 3$ hops, the routing position is recomputed from the updated token state (\emph{semantic re-routing}).

\Paragraph{Rank-1 experts as monosemantic units.}
Each expert stores one $W_\text{down} \in \mathbb{R}^{1 \times \dmodel}$ (input direction) and one $W_\text{up} \in \mathbb{R}^{\dmodel \times 1}$ (output direction), connected by SiLU nonlinearity.
This rank-1 structure means each expert reads along \emph{one} direction in activation space and writes along \emph{one} direction---an architectural analogue of sparse autoencoder (SAE) decoder columns \cite{bricken2023monosemanticity}, but trained end-to-end rather than post-hoc, eliminating the reconstruction-fidelity gap.
The SiLU gate ensures these are genuine computational units: $\cos(W_\text{down}, W_\text{up})$ averages 0.157 across all 8,192 experts, with 62.3\% near-orthogonal ($|\cos| < 0.2$) and 0\% identity-like ($|\cos| > 0.8$).

\Paragraph{Cosine routing as geometric substrate.}
Unlike standard linear routers ($W_r \in \mathbb{R}^{\dmodel \times M}$), cosine routing places tokens and experts in a shared metric space.
This geometric structure enables direct inspection: an expert's centroid position encodes \emph{what} it specializes in, and the distance between a token's position and an expert's centroid encodes \emph{how strongly} it will be activated.
This transparency is the key enabler of the interventions in this paper.

\Paragraph{Experimental setup.}
All results use the Marathon configuration: $\dmodel = 1024$, 16 heads, 8 layers, $M = 1024$ experts/layer, 50K training steps on WikiText-103 (1.64B tokens, $\sim$14 epochs).
Total parameters: 76--84M depending on routing variant.
Converged Wide 1$\times$12 checkpoint (PPL 33.93) and Deep 3$\times$4 checkpoint (PPL 34.62) are used for all analyses unless noted otherwise.
See \cite{equifinality2026} for full training details, convergence validation, and multi-seed statistics.

% ============================================================
% 3. LOGIT LENS
% ============================================================
\section{Logit Lens: Experts as Semantic Stepping Stones}
\label{sec:logit_lens}

We apply the logit lens technique \cite{nostalgebraist2020logit, geva2022transformer}: probing intermediate representations through the unembedding matrix to measure $P(\text{target token})$ at each point in the multi-hop trajectory.
We use the 36M-scale model in the underfitting regime (PPL $\sim$300--320) for this analysis: higher per-token entropy makes hop-by-hop probability changes more visible, providing clearer mechanistic characterization.
We intercept $h_\text{current} = x + h_\text{accum}$ before and after each hop, apply the final LayerNorm and LM head, and record the probability assigned to the ground-truth next token.
All remaining analyses in this paper use the fully converged 76--84M checkpoints.

\begin{figure}[t]
    \centering
    \includegraphics[width=\textwidth]{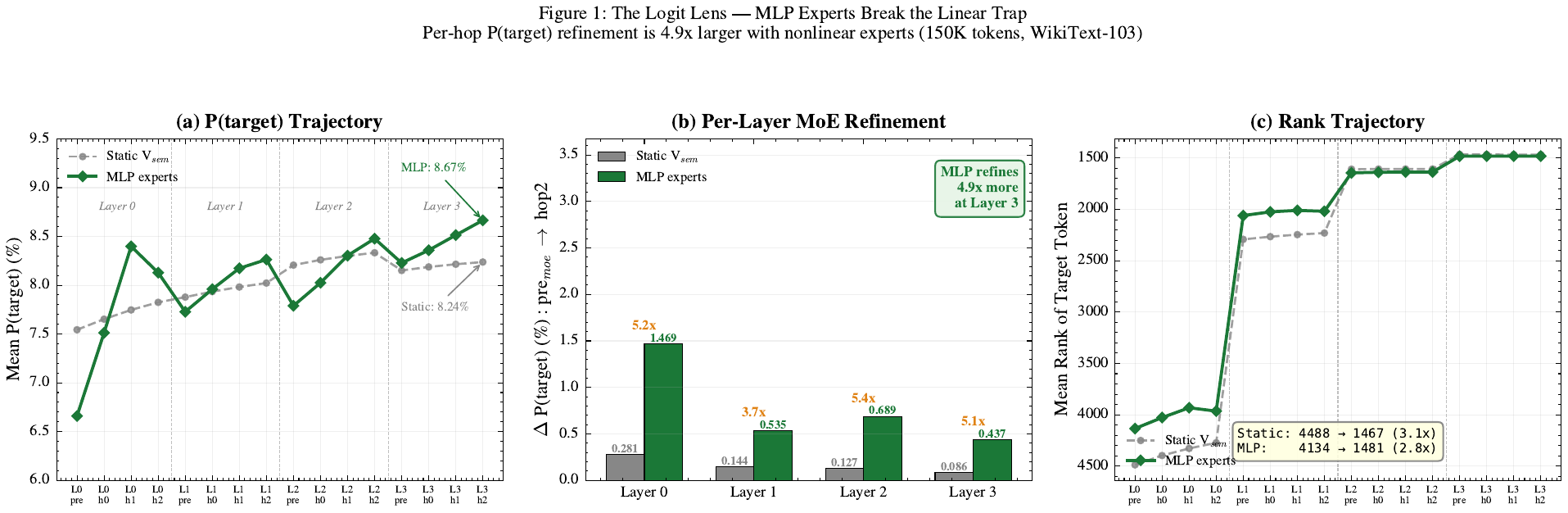}
    \caption{\textbf{The Logit Lens.} (a)~$P(\text{target})$ rises monotonically with each hop for both static and MLP experts, with MLP showing $4.8\times$ larger refinement. (b)~Per-layer MoE refinement: MLP experts contribute $+4.3\%$ gain at Layer~3 vs $+0.9\%$ for static. (c)~Mean rank of target token decreases with hops, confirming directional progress.}
    \label{fig:logit_lens}
\end{figure}

$P(\text{target})$ increases monotonically at every hop in every layer (Figure~\ref{fig:logit_lens}).
At Layer~3 (the final transformer layer), the total MoE contribution increases $P(\text{target})$ by $+4.3\%$ with MLP experts vs $+0.9\%$ with static vectors---a $4.8\times$ amplification.
The mean rank of the target token decreases with each hop, confirming directional progress rather than random perturbation.
However, as the convergence analysis reveals \cite{equifinality2026}, multi-hop updates are collinear ($\cos = 0.805$), implementing repeated amplification in the same direction rather than compositional reasoning.
The logit lens confirms hops do useful work; the echo chamber analysis in the companion paper explains why they are replaceable by simpler mechanisms.

% ============================================================
% 4. SEMANTIC DICTIONARY
% ============================================================
\section{Semantic Dictionary: Decoding Expert Functions}
\label{sec:dictionary}

Following the approach of Geva et al.\ \cite{geva2022transformer} for dense FFN layers, we project each expert's $W_\text{up}$ vector through the unembedding matrix to decode what it writes to the residual stream in vocabulary space.
This is a straightforward application of existing techniques to MoE experts, not a novel method.

\begin{table}[t]
\centering
\caption{Semantic Dictionary: representative expert archetypes (converged 50K checkpoint).}
\label{tab:dictionary}
\begin{tabular}{llll}
\toprule
Layer & Expert & Top Decoded Words & Archetype \\
\midrule
L7 & E760 & before, after, prior, earlier & Temporal sequence \\
L3 & E71 & six, three, four, five, eight & Cardinal numbers \\
L7 & E193 & Moreover, Indeed, Furthermore & Discourse connectives \\
L6 & E224 & trouble, pain, disease, surgery & Medical/adversity \\
L6 & E88 & -ing, -ed, -ly, -tion, -ment & Morphological suffixes \\
L5 & E733 & London, Paris, Berlin, Rome & European capitals \\
\midrule
L0 & E12 & the, comma, period, of, and & Garbage collector \\
L0 & E999 & \#\#, @@, [[, ]], \{\{\}\} & Tokenizer artifacts \\
\bottomrule
\end{tabular}
\end{table}

Representative archetypes are shown in Table~\ref{tab:dictionary}.
Based on manual inspection of top-10 decoded vocabulary words for all 8{,}192 experts, approximately 15\% are crystal-clear monosemantic specialists (top-10 words fall within a single coherent category), 8\% are functional transformers (reading one concept, writing another---\eg, verb stems $\to$ morphological suffixes), 12\% are garbage collectors absorbing high-frequency tokens and tokenizer artifacts, and the remaining 65\% are moderately polysemantic \cite{elhage2022toy}.
Early layers (L0--L2) handle syntax; late layers (L6--L7) encode the most specific semantic functions---mirroring findings in dense Transformer circuits.

\Paragraph{Automated category discovery.}
Beyond manual inspection, we apply HDBSCAN clustering on decoded expert centroids (top-10 token embedding centroids for all 8{,}192 experts) to discover categories that manual labeling might miss.
The algorithm identifies 20 clusters beyond the manually curated categories, including small integers (cluster of 36 experts, coherence 0.47), auxiliary verbs (13 experts, coherence 0.36), modern years (18 experts, coherence 0.57), and creation verbs (7 experts, coherence 0.48).
Notably, the model fragments numeric processing across 5 distinct clusters (small integers 1--9, mid-range 11--18, larger 20--30, ordinal words, ordinal suffixes), suggesting fine-grained numeric specialization at the expert level.
Of the novel clusters, two---small integers and auxiliary verbs---are validated as steerable in Section~\ref{sec:robustness}, bringing the total to 10 causally validated semantic categories.

% ============================================================
% 5. FREQUENCY CONTROL
% ============================================================
\section{Frequency-to-Syntax Gradient}
\label{sec:frequency}

A natural objection is that the early-layer syntactic specialization merely reflects Zipf's-law frequency clustering: function words are the ${\sim}200$ most frequent tokens, so gradient magnitude alone could explain their routing separation.
We control for this with a three-way token split: function words (24.7\% of tokens), the 50 most frequent \emph{content} words (\eg, ``time'', ``Australian'', ``city''; 2.5\%), and low-frequency content words (38.4\%).
If the router learned frequency, high-frequency content words should cluster with function words in routing space; if it learned syntax, they should cluster with other content words.

We compare silhouette scores under syntax grouping (function vs.\ all content) and frequency grouping (high-frequency vs.\ low-frequency) with permutation null (200 permutations) and bootstrap confidence intervals (500 resamples).
Syntax grouping significantly outperforms frequency grouping at all 8 layers ($\Delta > 0$, all 95\% CIs exclude zero, $p < 0.001$), with a Spearman depth trend of $\rho = 0.81$ ($p = 0.015$)---syntax dominance \emph{increases} with depth.
As a gold-standard control, we test 15 matched-frequency pairs (function/content words with $< 2\times$ corpus frequency ratio): even at identical frequency, the router separates them by syntactic class at every layer (silhouette 0.18--0.46, all $p < 0.001$).

We further distinguish \emph{grammar-enriched} experts (odds ratio OR $> 1.5$, FDR $< 0.05$; 30--40\% per layer) from \emph{grammar-specialist} experts (OR $> 10$, FDR $< 0.05$; 77 at L0).
The top L0 specialists process $> 98\%$ function words and decode to coherent grammatical categories: plural auxiliaries (``were'', ``are'', ``both''), conjunctions (``or'', ``and'', ``but''), and pronouns (``Her'', ``She'', ``Our'').
Expert~52 groups ``both'' and ``each'' (low frequency) with ``were'' and ``are'' (high frequency)---syntactic role, not corpus frequency, is the organizing principle.
This gradient from coarse frequency sensitivity in early layers to fine-grained syntactic specialization in deeper layers mirrors established findings in dense Transformers \cite{jawahar2019does, tenney2019bert}.

% ============================================================
% 6. POLYSEMY BRANCHING
% ============================================================
\section{Polysemy Branching}
\label{sec:polysemy}

\begin{figure}[t]
    \centering
    \includegraphics[width=0.85\textwidth]{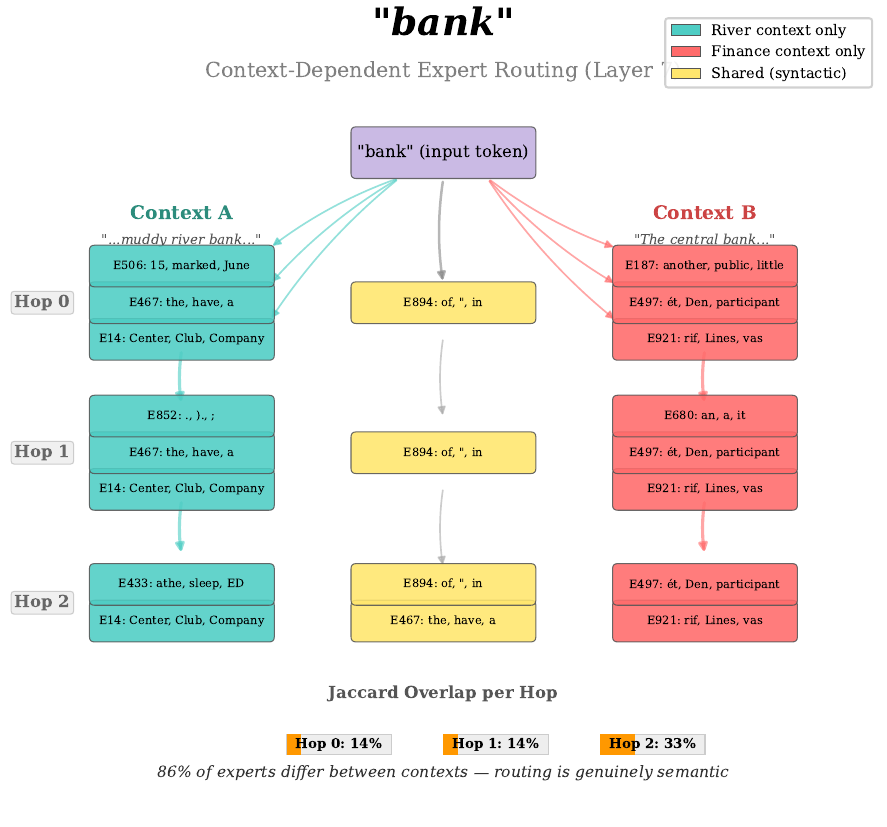}
    \caption{\textbf{Polysemy Branching.} The token ``bank'' routes through 86\% different experts depending on river vs.\ finance context. Shared experts (yellow) handle syntactic obligations; unique experts (teal/red) encode context-specific semantics.}
    \label{fig:polysemy}
\end{figure}

Feeding the identical token ``bank'' in two contexts (``muddy river bank'' vs ``The central bank'') produces expert trajectories that diverge by 86\% (Jaccard overlap: 14.3\% at Hop~0, 14.3\% at Hop~1, 33.3\% at Hop~2).
The few shared experts decode as syntactic-glue specialists (determiners, function words), consistent with the grammar-specialist experts identified in Section~\ref{sec:frequency}.
This demonstrates that routing is \textbf{genuinely semantic, not positional}---the same input token activates completely different experts depending on attended context (Figure~\ref{fig:polysemy}).

% ============================================================
% 7. ROUTING STATISTICS
% ============================================================
\section{Quantitative Routing Statistics}
\label{sec:routing_stats}

\begin{figure}[t]
    \centering
    \includegraphics[width=\textwidth]{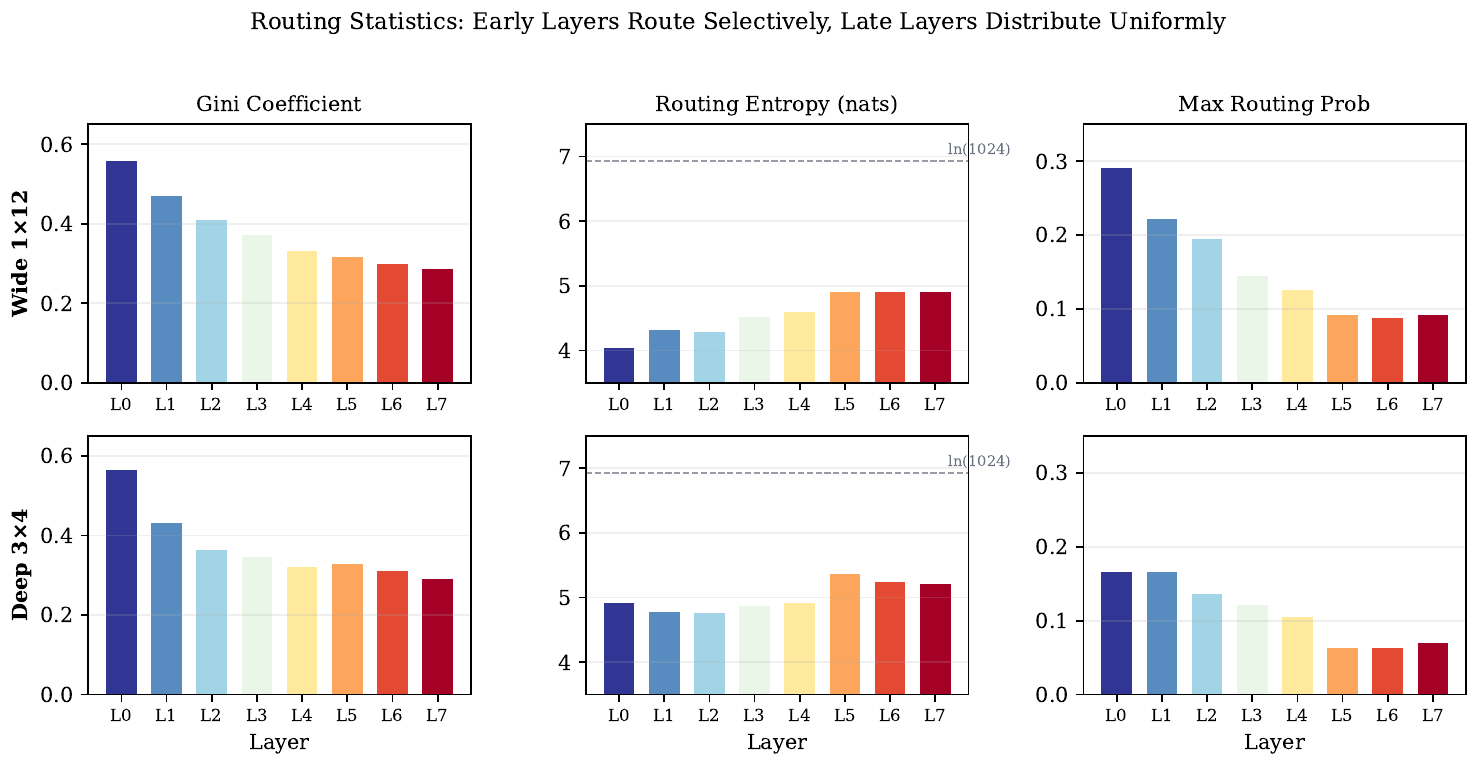}
    \caption{\textbf{Routing Statistics by Layer.} Both Wide and Deep architectures exhibit a clear gradient: early layers (L0--L2) route selectively (high Gini, low entropy, high max probability), while late layers (L5--L7) distribute near-uniformly across experts. This pattern holds regardless of routing topology.}
    \label{fig:routing_stats}
\end{figure}

Having shown that routing depends on semantic context (Section~\ref{sec:polysemy}), we next quantify how selective routing is across layers.
We compute formal routing metrics over 50 validation batches.
Of 8{,}192 total experts, only 3 are dead (zero activations)---near-perfect utilization.
Routing exhibits a clear layer-wise gradient:
early layers route selectively (L0: utilization Gini $= 0.56$, mean max routing probability $= 0.29$),
while late layers distribute more uniformly (L7: Gini $= 0.29$, max prob $= 0.09$, routing entropy 4.92 nats approaching the uniform maximum of $\ln 1024 = 6.93$).
This quantifies the qualitative observation that early layers perform coarse semantic categorization while late layers combine broader expert ensembles (Figure~\ref{fig:routing_stats}).

% ============================================================
% 8. CAUSAL INTERVENTIONS
% ============================================================
\section{Causal Interventions}
\label{sec:interventions}

The preceding sections establish that experts develop interpretable specializations and that routing organizes tokens semantically.
However, these are \emph{correlational} observations.
To establish \emph{causation}---that expert labels predict output behavior---we perform four inference-time interventions on converged checkpoints.

\subsection{Targeted Knockout}

We discover which experts are actually routed to for a given prompt, then zero the top-weighted expert's $W_\text{up}$ and $W_\text{down}$ at each layer individually.
Single-expert knockout causes KL divergences up to \textbf{0.468} (L6-E420 for medical context) and \textbf{0.289} (L5-E468 for temporal context)---comparable to removing an entire attention head in standard Transformers.
For the prompt \emph{``The patient was diagnosed with a serious''}, knocking out L6-E420 shifts the distribution: P(brain) drops from 15.6\% to 3.9\% ($-75\%$).
Averaging across prompts, L0 (avg KL=0.232) and L6 (avg KL=0.162) are the most causally important layers, confirming L6 as the primary semantic shaping layer.

\subsection{Routing Steering}
\label{sec:steering}

We bias the routing position toward a target expert's centroid ($\text{pos}_\text{steered} = \text{normalize}(\text{pos} + \lambda \cdot c_\text{target})$) across 4 semantic categories.
Steering produces large, dose-dependent shifts:
on category-biased prompts, temporal $+191\%$ ($\lambda = 1.0$), cardinal $+63\%$, geographic $+41\%$, and discourse $+7\%$.
Across 44 diverse prompts (category-relevant, neutral, and adversarial), temporal steering has a median effect of $+321\%$ (IQR $[+124\%, +549\%]$, 98\% positive).
At $\lambda = 1.0$, temporal steering shifts P(temporal) from 0.17 to 0.49---nearly half the next-token probability mass lands on temporal words---and NLL \emph{improves} by $-15\%$, indicating genuine signal amplification, not noise injection.

\subsection{Expert Suppression}

We block routing to clusters of $N$ experts (setting their similarity to $-\infty$) and compare against random suppression of the same size.
Suppressing a single expert produces large, selective output shifts (Table~\ref{tab:suppression}):

\begin{table}[h]
\centering
\caption{Expert suppression ($N{=}1$) on category-relevant prompts: blocking one correctly-identified expert produces large, selective output shifts. Random suppression of the same size has zero effect.}
\label{tab:suppression}
\begin{tabular}{lcccc}
\toprule
Category & $\Delta P$ (targeted) & $\Delta P$ (random) & KL (targeted) & KL (random) \\
\midrule
Temporal (L4-E80)    & $\mathbf{-33.7\%}$ & $+0.0\%$ & $\mathbf{0.139}$ & 0.000 \\
Geographic (L4-E685) & $\mathbf{-23.2\%}$ & $+0.0\%$ & $\mathbf{0.512}$ & 0.000 \\
Cardinal (L7-E1004)  & $-0.0\%$ & $+0.0\%$ & 0.000 & 0.000 \\
Discourse (L7-E525)  & $-0.4\%$ & $-0.0\%$ & 0.000 & 0.000 \\
\bottomrule
\end{tabular}
\end{table}

Suppressing 1 expert accounts for the entire cluster effect: adding 4--99 neighbors produces no additional change ($-33.7\%$ at $N{=}1$ vs.\ $-33.4\%$ at $N{=}100$).
Categories with low expert selection rates (cardinal, discourse) show near-zero effects because the target expert is rarely in the top-$k$ for test prompts---consistent with content-addressed routing.

\subsection{Expert Surgery}

We directly rewrite an expert's $\mathbf{W}_\text{up}$ vector.
Replacing the geographic expert's output vector halves P(geographic words) on category-relevant prompts ($\Delta P = -45.8\%$) with zero change on control prompts---confirming that individual experts are causally responsible for specific semantic content.

\subsection{Read/Write Direction Analysis}

For all 8{,}192 experts, we compute $\cos(W_\text{down}, W_\text{up})$ to test whether experts are trivial copy-paste operations.
The overall mean cosine is \textbf{0.157}; \textbf{0.0\%} of experts are identity-like ($|\cos| > 0.8$) while \textbf{62.3\%} are near-orthogonal ($|\cos| < 0.2$).
Experts genuinely read one concept and write a different, orthogonal concept through SiLU nonlinearity---they are computational units, not lookup tables.

% ============================================================
% 9. STEERING ROBUSTNESS AND GENERALITY
% ============================================================
\section{Steering Robustness and Generality}
\label{sec:robustness}

The $+191\%$ temporal effect above was measured on category-relevant prompts chosen to activate the target expert.
To evaluate robustness, we test the same steering intervention ($\lambda = 1.0$) on 44 temporal and 39 geographic prompts spanning category-relevant, neutral, and adversarial contexts (\eg, ``The recipe calls for'', ``The chemical formula is'').
Temporal steering remains positive on 98\% of prompts (median $+321\%$, IQR $[+124\%, +549\%]$), with only 1/44 prompts showing a negative shift ($-5.5\%$).
Geographic steering is positive on 90\% (median $+56\%$, IQR $[+19\%, +151\%]$), with 4/39 negative.

To test generality beyond temporal and geographic categories, we use the Semantic Dictionary to automatically discover experts for six additional categories.
For each category, we project all experts' $\mathbf{W}_\text{up}$ through the unembedding matrix, filter hub experts (those with uniformly high similarity to random vocabulary), and validate that decoded vocabulary overlaps with category seed words.
Steering the discovered experts at $\lambda = 1.0$ produces positive $\Delta P$ for all six categories:
emotional $+71\%$ (L7-E842: ``upset, exhausted, weakened''),
financial $+53\%$ (L5-E478: ``profit, cost''),
military $+20\%$ (L0-E410: ``force, forces, movement''),
legal $+19\%$ (L0-E144: ``court, case''),
scientific $+18\%$ (L0-E873: ``study, theory''),
and social $+3\%$ (L7-E888: ``family, team, staff'').
Automated clustering via HDBSCAN (Section~\ref{sec:dictionary}) discovers two additional steerable categories:
small integers $+84\%$ (L3-E14, cluster of 36 experts decoding digits 1--9; 90\% positive across 20 prompts)
and auxiliary verbs $+14\%$ (L6-E6, cluster of 13 experts decoding ``was'', ``had'', ``were''; 75\% positive across 20 prompts).
Small integers is among the strongest steering effects observed in any experiment, comparable to the manually identified temporal category.
Across all 10 categories (2 original, 6 dictionary-discovered, 2 clustering-discovered), steering effects range from $+3\%$ to $+453\%$.

% ============================================================
% 10. COMPOSITIONAL STEERING
% ============================================================
\section{Compositional Steering}
\label{sec:composition}

Compositional steering---activating two experts simultaneously---reveals a structural constraint: experts \emph{within the same layer} interfere (temporal+geographic, both L4: solo $+567\%$/$+44\%$ $\to$ composed $+41\%$/$+35\%$), while experts \emph{across layers} compose nearly additively (temporal L4 + cardinal L7: solo $+567\%$/$+19\%$ $\to$ composed $+513\%$/$+18\%$).
Cross-talk is low in both cases ($<10\%$), suggesting expert representations are largely orthogonal; the interference arises from centroid-addition steering modifying routing for \emph{all} experts in the target layer.

This reveals an important structural property: compositional expert control operates \emph{across} layers, not \emph{within} them.
Same-layer interference is a consequence of the routing mechanism (centroid addition perturbs all cosine similarities in that layer), not of expert representation overlap.
This has practical implications: multi-category steering should target experts in different layers for near-additive composition.

% ============================================================
% 11. CROSS-ROUTING COMPARISON
% ============================================================
\section{Cross-Routing Controllability}
\label{sec:cross_routing}

Are these interventions unique to cosine routing, or do linear-router experts exhibit similar controllability?
We apply the same three tests to the linear router checkpoint (PPL 32.76), discovering target experts via activation analysis (since there are no geometric centroids to inspect).

Both routing modes support steering: cosine temporal steering reaches $+107\%$; linear geographic steering reaches $+293\%$.
However, the \emph{mechanism} differs fundamentally.
Cosine steering biases a geometric position toward a target centroid---an interpretable operation in routing space.
Linear steering adds a scalar bias to one logit dimension---effective but opaque.
The advantage of cosine routing is not exclusive controllability but \textbf{geometric transparency}: experts can be discovered, inspected, and manipulated via their positions in routing space without running activation probes.
Linear routing requires running category-specific prompts through the model to identify relevant experts; cosine routing lets you read them from the centroid matrix.

% ============================================================
% 12. DISCUSSION
% ============================================================
\section{Discussion}
\label{sec:discussion}

\Paragraph{Scale limitations.}
All results are on WikiText-103 at 76--84M parameters.
While the equifinality finding from \cite{equifinality2026} is supported by concurrent scaling law work that treats routing as fixed infrastructure \cite{zhao2025comprehensive, tian2025leverage}, the controllability effects documented here---steering magnitudes, suppression selectivity, and compositional behavior---may change at billion-parameter scale.
In particular, larger expert pools with higher rank ($r > 1$) could exhibit richer polysemanticity and different composition dynamics.
Validating these interventions at scale is the most important follow-up.

\Paragraph{Relationship to SAEs and post-hoc interpretability.}
Our rank-1 MoE experts are structurally analogous to SAE decoder columns \cite{bricken2023monosemanticity}---both are directions in activation space that are sparsely activated.
However, experts are trained end-to-end, eliminating the reconstruction-fidelity gap and sidestepping recently exposed SAE limitations (non-canonical features \cite{leask2025saes}, absorption \cite{chanin2024absorption}, predictable dark matter \cite{engels2024dark}; see Section~\ref{sec:related} for details).
The trade-off is generality: experts are tied to the MoE architecture, while SAEs can be applied to any model.

\Paragraph{Comparison with dense-model steering.}
Dense-model steering methods (ITI \cite{li2024inference}, Representation Engineering \cite{zou2023representation}, Activation Addition \cite{turner2023activation}, ROME \cite{meng2022locating}) all require contrastive datasets, linear probes, or causal tracing to identify targets (see Section~\ref{sec:related}).
Our approach sidesteps this pipeline: rank-1 experts are monosemantic by construction, and cosine centroids reveal their function directly.
Concurrently, SteerMoE \cite{fayyaz2025steermoe} demonstrates expert (de)activation for safety steering via behavioral frequency; our geometric approach complements their behavioral one.

\Paragraph{Numeric fragmentation.}
Automated expert decoding reveals that numeric processing is distributed across 5 specialized clusters rather than a single ``number'' category: small integers (1--9), mid-range integers (11--18), larger integers (20--30), ordinal words (``third'', ``fourth''), and ordinal suffixes (``1st'', ``2nd'').
This fragmentation suggests the model develops fine-grained numeric representations at the expert level, with each cluster capturing a distinct numeric register.
The small integers cluster (36 experts, coherence 0.47) produces the strongest steering effect ($+84\%$) of any automatically discovered category, indicating that single-digit processing is a particularly concentrated expert function.

\Paragraph{Practical implications.}
The zero-overhead nature of our interventions---no additional training, no probing datasets, no post-hoc analysis---makes expert-level control practical for deployment.
The compositional constraint (cross-layer but not within-layer) provides a concrete guideline for multi-category steering.
However, we caution that the safety-adjacent categories (military, legal) show smaller effects ($+19\%$--$+20\%$) than semantic categories ($+321\%$ temporal), suggesting that safety-relevant content may be more distributed across experts.

% ============================================================
% 13. RELATED WORK
% ============================================================
\section{Related Work}
\label{sec:related}

\Paragraph{Sparse Mixture of Experts.}
Sparse MoE was introduced by Shazeer et al.\ \cite{shazeer2017outrageously} with top-2 gating.
Switch Transformer \cite{fedus2022switch} simplified to top-1 with load balancing.
GShard \cite{lepikhin2021gshard} scaled to 600B parameters.
Our companion paper \cite{equifinality2026} provides the full architectural context and equifinality results that motivate this work.

\Paragraph{Logit lens and mechanistic interpretability.}
The logit lens \cite{nostalgebraist2020logit} probes intermediate representations through unembedding.
Geva et al.\ \cite{geva2022transformer} showed FFN layers write vocabulary-space predictions to the residual stream.
We extend this to multi-hop MoE, showing monotonic progress across hops and using unembedding projections to decode individual expert functions.

\Paragraph{Mechanistic interpretability and monosemanticity.}
Elhage et al.\ \cite{elhage2022toy} introduced the superposition hypothesis: neural networks pack more features than dimensions by exploiting near-orthogonal directions.
Sparse autoencoders (SAEs) decompose these superposed activations into overcomplete dictionaries of monosemantic features, scaling from 1-layer toy models \cite{bricken2023monosemanticity} to 34M features on Claude 3 Sonnet \cite{templeton2024scaling} and 16M features on GPT-4 \cite{gao2024scaling}.
However, recent work has exposed fundamental SAE limitations: trained SAEs do not find canonical features, with only 30--40\% of features shared across runs \cite{leask2025saes}; L1 sparsity causes feature absorption \cite{chanin2024absorption}; and $>$90\% of reconstruction error is linearly predictable \cite{engels2024dark}.
Our rank-1 MoE experts are structurally analogous to SAE decoder columns but trained end-to-end, eliminating the reconstruction-fidelity gap.
Independently, Chen et al.\ \cite{chen2025sparsity} showed that MoE models exhibit less superposition than dense models with equivalent parameters.
MONET \cite{park2025monet} pursues the same goal at larger scale (262K monosemantic experts via product key composition), though without geometric routing transparency.
Our Semantic Dictionary mirrors the logit lens on SAE features \cite{bloom2024logitlens}: projecting expert $\mathbf{W}_\text{up}$ through the unembedding reveals expert functions directly, at zero cost and with no approximation gap.

\Paragraph{Inference-time steering and model editing.}
Inference-Time Intervention \cite{li2024inference} shifts attention head activations along learned truthfulness directions.
Representation Engineering \cite{zou2023representation} and Activation Addition \cite{turner2023activation} extract steering vectors from contrastive prompt pairs.
ROME \cite{meng2022locating} performs rank-one weight edits on MLP layers identified via causal tracing.
These methods operate on dense models and require contrastive datasets, linear probes, or causal tracing.
Our approach exploits MoE expert structure: rank-1 experts are monosemantic by construction, and cosine routing centroids reveal their semantic function directly.
We combine three intervention types (steering, suppression, surgery) on the same routing-identified targets, yielding large effect sizes (median $+321\%$) generalizing to 10 semantic categories.
Concurrently, SteerMoE \cite{fayyaz2025steermoe} demonstrates expert (de)activation for safety steering using behavioral activation frequency.
Recent work on SAE-based steering \cite{arad2025saes} shows that feature selection is critical; our rank-1 experts avoid this misalignment since each expert directly writes its direction to the residual stream.

\Paragraph{Post-hoc and model-agnostic explainability.}
Model-agnostic methods such as LIME \cite{ribeiro2016lime} and SHAP \cite{lundberg2017shap} provide feature attribution via local surrogate models.
Recent extensions to LLMs include CELL \cite{luss2024cell} (contrastive explanations) and gSMILE \cite{dehghani2025gsmile} (token-level attribution via Wasserstein distance).
These methods work with black-box API access but carry limitations: surrogate approximations may not faithfully represent actual computation, explanations are local with no global view, and computational cost is high.
Our approach trades generality for faithfulness: cosine routing makes expert selection directly inspectable, expert surgery provides causal validation, and the Semantic Dictionary yields global expert-level understanding at zero additional cost.

\Paragraph{Expert specialization in MoE.}
DeepSeek-MoE \cite{dai2024deepseekmoe} documents expert specialization at scale but does not perform causal validation.
Wang et al.\ \cite{wang2026illusion} find domain-invariant ``standing committee'' experts but provide only observational evidence.
Our causal interventions complement these findings by demonstrating that expert specializations are not merely correlational but causally responsible for specific output distributions.

% ============================================================
% 14. CONCLUSION
% ============================================================
\section{Conclusion}
\label{sec:conclusion}

We have shown that individual rank-1 MoE experts develop interpretable semantic specializations that are causally responsible for specific output distributions.
Projecting expert output vectors through the unembedding matrix reveals expert functions spanning 10 semantic categories, routing organizes tokens by syntactic class with a depth gradient, and three causal interventions---steering, suppression, and surgery---confirm that these labels predict model behavior, not merely describe it.

These results complement the equifinality finding from \cite{equifinality2026}: routing \emph{topology} is quality-neutral, but expert \emph{identity} is causally meaningful.
The key practical insight is that cosine routing provides geometric transparency---expert functions are readable directly from centroids without running activation probes---though our cross-routing comparison shows linear routers support qualitatively similar causal effects.
The most important open question is whether these findings extend to billion-parameter MoE models with higher-rank experts and broader semantic coverage.

\Paragraph{Limitations.}
All results are on WikiText-103 at 76--84M parameters with rank-1 experts; we do not claim generalization to billion-parameter scale, diverse corpora, full-rank expert architectures, or production MoE systems without further validation.
Causal interventions were validated on 10 semantic categories (8 manually identified, 2 discovered via HDBSCAN clustering), and automated clustering reveals at least 20 coherent expert clusters---though 73.5\% of experts remain unclustered, and only 7 of 16 novel clusters were steering-tested.
Broader coverage may reveal experts with less clean specialization or weaker causal effects.
Safety-adjacent steering categories show smaller effects than semantic categories, and the compositional constraint (cross-layer only) limits multi-dimensional control within single layers.

\Paragraph{Implication.}
MoE expert-level specialization is a first-class interpretability primitive: architecturally monosemantic, causally validated, and controllable at inference time with zero overhead.
For the interpretability community, these results suggest that MoE architectures offer a natural decomposition into semantically meaningful units---an alternative to post-hoc dictionary learning that avoids the reconstruction-fidelity trade-off inherent in sparse autoencoders.

% ============================================================
% REFERENCES
% ============================================================
\bibliographystyle{unsrt}
\bibliography{references}

@article{shazeer2017outrageously,
  title={Outrageously Large Neural Networks: The Sparsely-Gated Mixture-of-Experts Layer},
  author={Shazeer, Noam and Mirhoseini, Azalia and Maziarz, Krzysztof and Davis, Andy and Le, Quoc and Hinton, Geoffrey and Dean, Jeff},
  journal={arXiv preprint arXiv:1701.06538},
  year={2017}
}

@article{fedus2022switch,
  title={Switch Transformers: Scaling to Trillion Parameter Models with Simple and Efficient Sparsity},
  author={Fedus, William and Zoph, Barret and Shazeer, Noam},
  journal={Journal of Machine Learning Research},
  volume={23},
  number={120},
  pages={1--39},
  year={2022}
}

@inproceedings{lepikhin2021gshard,
  title={{GShard}: Scaling Giant Models with Conditional Computation and Automatic Sharding},
  author={Lepikhin, Dmitry and Lee, HyoukJoong and Xu, Yuanzhong and Chen, Dehao and Firat, Orhan and Huang, Yanping and Krikun, Maxim and Shazeer, Noam and Chen, Zhifeng},
  booktitle={International Conference on Learning Representations},
  year={2021}
}

@article{nostalgebraist2020logit,
  title={interpreting {GPT}: the logit lens},
  author={nostalgebraist},
  journal={LessWrong},
  year={2020},
  url={https://www.lesswrong.com/posts/AcKRB8wDpdaN6v6ru/interpreting-gpt-the-logit-lens}
}

@inproceedings{geva2022transformer,
  title={Transformer Feed-Forward Layers Build Predictions by Promoting Concepts in the Vocabulary Space},
  author={Geva, Mor and Caciularu, Avi and Wang, Kevin and Goldberg, Yoav},
  booktitle={Proceedings of the 2022 Conference on Empirical Methods in Natural Language Processing},
  year={2022}
}

@article{elhage2022toy,
  title={Toy Models of Superposition},
  author={Elhage, Nelson and Hume, Tristan and Olsson, Catherine and Schiefer, Nicholas and Henighan, Tom and Kravec, Shauna and Hatfield-Dodds, Zac and Lasenby, Robert and Drain, Dawn and Chen, Carol and others},
  journal={Transformer Circuits Thread},
  year={2022},
  url={https://transformer-circuits.pub/2022/toy_model/index.html}
}

@article{bricken2023monosemanticity,
  title={Towards Monosemanticity: Decomposing Language Models With Dictionary Learning},
  author={Bricken, Trenton and Templeton, Adly and Batson, Joshua and Chen, Brian and Jermyn, Adam and Conerly, Tom and Turner, Nick and Anil, Cem and Denison, Carson and Askell, Amanda and others},
  journal={Transformer Circuits Thread},
  year={2023},
  url={https://transformer-circuits.pub/2023/monosemantic-features/index.html}
}

@article{templeton2024scaling,
  title={Scaling Monosemanticity: Extracting Interpretable Features from {Claude} 3 {Sonnet}},
  author={Templeton, Adly and Conerly, Tom and Marcus, Jonathan and Lindsey, Jack and Bricken, Trenton and Chen, Brian and Pearce, Adam and Citro, Craig and Ameisen, Emmanuel and Jones, Andy and Cunningham, Hoagy and Turner, Nicholas L and McDougall, Callum and MacDiarmid, Monte and Freeman, C Daniel and Sumers, Theodore R and Rees, Edward and Batson, Joshua and Jermyn, Adam and Carter, Shan and Olah, Chris and Henighan, Tom},
  journal={Transformer Circuits Thread},
  year={2024}
}

@article{gao2024scaling,
  title={Scaling and Evaluating Sparse Autoencoders},
  author={Gao, Leo and la Tour, Tom Dupr{\'e} and Tillman, Henk and Goh, Gabriel and Troll, Rajan and Radford, Alec and Sutskever, Ilya and Leike, Jan and Wu, Jeffrey},
  journal={arXiv preprint arXiv:2406.04093},
  year={2024}
}

@article{leask2025saes,
  title={Sparse Autoencoders Do Not Find Canonical Units of Analysis},
  author={Leask, Patrick and Mendel, Joshua and Boettiger, Stepan and Mulligan, Nikhil and others},
  journal={Proceedings of the International Conference on Learning Representations},
  year={2025}
}

@article{chanin2024absorption,
  title={A is for Absorption: Studying Feature Splitting and Absorption in Sparse Autoencoders},
  author={Chanin, David and Wilken-Smith, James and Dulka, Tom{\'a}{\v{s}} and Bhatnagar, Hardik and Bloom, Joseph},
  journal={arXiv preprint arXiv:2409.14507},
  year={2024}
}

@article{engels2024dark,
  title={Decomposing The Dark Matter of Sparse Autoencoders},
  author={Engels, Joshua and Liao, Isaac and Tegmark, Max},
  journal={arXiv preprint arXiv:2410.14670},
  year={2024}
}

@article{chen2025sparsity,
  title={Sparsity and Superposition in Mixture of Experts},
  author={Chen, Tianyi and others},
  journal={arXiv preprint arXiv:2510.23671},
  year={2025}
}

@inproceedings{park2025monet,
  title={{MONET}: Mixture of Monosemantic Experts for Transformers},
  author={Park, Jungwoo and Ahn, Young Jin and Kim, Kee-Eung and Kang, Jaewoo},
  booktitle={Proceedings of the International Conference on Learning Representations},
  year={2025}
}

@article{bloom2024logitlens,
  title={Understanding {SAE} Features with the Logit Lens},
  author={Bloom, Joseph and Lin, Curt Tigges},
  journal={LessWrong / Alignment Forum},
  year={2024}
}

@article{zou2023representation,
  title={Representation Engineering: A Top-Down Approach to {AI} Transparency},
  author={Zou, Andy and Phan, Long and Chen, Sarah and Campbell, James and Guo, Phillip and Ren, Richard and Pan, Alexander and Yin, Xuwang and Mazeika, Mantas and Dombrowski, Ann-Kathrin and others},
  journal={arXiv preprint arXiv:2310.01405},
  year={2023}
}

@article{turner2023activation,
  title={Activation Addition: Steering Language Models Without Optimization},
  author={Turner, Alexander Matt and Thiergart, Lisa and Udell, David and Leech, Gavin and Mini, Ulisse and Pelrine, Monte},
  journal={arXiv preprint arXiv:2308.10248},
  year={2023}
}

@inproceedings{li2024inference,
  title={Inference-Time Intervention: Eliciting Truthful Answers from a Language Model},
  author={Li, Kenneth and Patel, Oam and Vi{\'e}gas, Fernanda and Pfister, Hanspeter and Wattenberg, Martin},
  booktitle={Advances in Neural Information Processing Systems},
  volume={36},
  year={2024}
}

@inproceedings{meng2022locating,
  title={Locating and Editing Factual Associations in {GPT}},
  author={Meng, Kevin and Bau, David and Andonian, Alex and Belinkov, Yonatan},
  booktitle={Advances in Neural Information Processing Systems},
  volume={35},
  year={2022}
}

@article{fayyaz2025steermoe,
  title={{SteerMoE}: Steering Mixture-of-Experts {LLMs} via Expert (De)Activation},
  author={Fayyaz, Mohsen and others},
  journal={arXiv preprint arXiv:2509.09660},
  year={2025}
}

@article{arad2025saes,
  title={{SAEs} Are Good for Steering -- If You Select the Right Features},
  author={Arad, Ido and Mueller, Aaron and Belinkov, Yonatan},
  journal={Proceedings of the 2025 Conference on Empirical Methods in Natural Language Processing},
  year={2025}
}

@inproceedings{ribeiro2016lime,
  title={``{Why} Should {I} Trust You?'': Explaining the Predictions of Any Classifier},
  author={Ribeiro, Marco Tulio and Singh, Sameer and Guestrin, Carlos},
  booktitle={Proceedings of the 22nd ACM SIGKDD International Conference on Knowledge Discovery and Data Mining},
  year={2016}
}

@inproceedings{lundberg2017shap,
  title={A Unified Approach to Interpreting Model Predictions},
  author={Lundberg, Scott M and Lee, Su-In},
  booktitle={Advances in Neural Information Processing Systems},
  volume={30},
  year={2017}
}

@article{luss2024cell,
  title={{CELL} your Model: Contrastive Explanations for Large Language Models},
  author={Luss, Ronny and Miehling, Erik and Dhurandhar, Amit},
  journal={arXiv preprint arXiv:2406.11785},
  year={2024}
}

@article{dehghani2025gsmile,
  title={Explaining Large Language Models with {gSMILE}},
  author={Dehghani, Zeinab and Akram, Mohammed Naveed and Aslansefat, Koorosh and Khan, Adil},
  journal={arXiv preprint arXiv:2505.21657},
  year={2025}
}

@inproceedings{jawahar2019does,
  title={What Does {BERT} Look at? An Analysis of {BERT}'s Attention},
  author={Jawahar, Ganesh and Sagot, Beno{\^i}t and Seddah, Djam{\'e}},
  booktitle={Proceedings of the 2019 ACL Workshop BlackboxNLP: Analyzing and Interpreting Neural Networks for NLP},
  pages={276--286},
  year={2019}
}

@article{tenney2019bert,
  title={{BERT} Rediscovers the Classical {NLP} Pipeline},
  author={Tenney, Ian and Das, Dipanjan and Pavlick, Ellie},
  journal={arXiv preprint arXiv:1905.05950},
  year={2019}
}

@article{dai2024deepseekmoe,
  title={DeepSeekMoE: Towards Ultimate Expert Specialization in Mixture-of-Experts Language Models},
  author={Dai, Damai and Deng, Chengqi and Zhao, Chenggang and Xu, R.X. and Gao, Huazuo and Chen, Deli and Li, Jiashi and Zeng, Wangding and Yu, Xingkai and Wu, Y. and others},
  journal={arXiv preprint arXiv:2401.06066},
  year={2024}
}

@article{jiang2024mixtral,
  title={Mixtral of Experts},
  author={Jiang, Albert Q and Sablayrolles, Alexandre and Roux, Antoine and Mensch, Arthur and Savary, Blanche and Bamford, Chris and Chaplot, Devendra Singh and de las Casas, Diego and Hanna, Emma Bou and Bressand, Florian and others},
  journal={arXiv preprint arXiv:2401.04088},
  year={2024}
}

@article{zhao2025comprehensive,
  title={Towards a Comprehensive Scaling Law of Mixture-of-Experts},
  author={Zhao, Guoliang and Fu, Yuhan and Li, Shuaipeng and others},
  journal={arXiv preprint arXiv:2509.23678},
  year={2025}
}

@article{tian2025leverage,
  title={Towards Greater Leverage: Scaling Laws for Efficient Mixture-of-Experts Language Models},
  author={Tian, Changxin and Chen, Kunlong and Liu, Jia and Liu, Ziqi and Zhang, Zhiqiang and Zhou, Jun},
  journal={arXiv preprint arXiv:2507.17702},
  year={2025}
}

@article{wang2026illusion,
  title={The Illusion of Specialization: Unveiling the Domain-Invariant ``Standing Committee'' in Mixture-of-Experts Models},
  author={Wang, Yan and Xu, Yitao and Shen, Nanhan and Su, Jinyan and Huang, Jimin and Zhu, Zining},
  journal={arXiv preprint arXiv:2601.03425},
  year={2026}
}

@article{equifinality2026,
  title={Equifinality in Mixture of Experts: Routing Topology Does Not Determine Language Modeling Quality},
  author={Ternovtsii, Ivan and Bilak, Yurii},
  journal={arXiv preprint},
  year={2026},
}

\end{document}